\documentclass[10pt,twocolumn,letterpaper]{article}

\usepackage{cvpr}
\usepackage{times}
\usepackage{epsfig}
\usepackage{graphicx}
\usepackage{amsmath}
\usepackage{amssymb}
\usepackage{subfigure}

\graphicspath{{./graphics/}}
\usepackage{animate}
\usepackage{tabularx}
\usepackage{booktabs}
\usepackage{caption}
\usepackage{tikz}
\usepackage{pgfplots}
\usepackage{pgfplotstable}
\usepackage[skins]{tcolorbox}
\usepackage{standalone}
\usepackage{bm}
\usepackage{mleftright}
\usepackage{nicefrac}
\usepackage{textcomp}
\usepackage[normalem]{ulem}
\usepackage[shortcuts]{extdash}
\newlength{\itemwidth}

\newcolumntype{Y}{>{\centering\arraybackslash}X}
\newcolumntype{P}[1]{>{\centering\arraybackslash}p{#1}}

\usetikzlibrary{calc}
\usetikzlibrary{tikzmark}
\pgfplotsset{compat=newest}

\definecolor{EE7F0E}{RGB}{238,127,14}
\definecolor{619D47}{RGB}{97,157,71}
\definecolor{3787CF}{RGB}{55,135,207}
\definecolor{DBDC4A}{RGB}{219,220,74}

\pgfplotscreateplotcyclelist{custompalette}{
    {EE7F0E!90!black,fill=EE7F0E},
    {619D47!90!black,fill=619D47},
    {3787CF!90!black,fill=3787CF},
    {DBDC4A!90!black,fill=DBDC4A}
}
\pgfplotscreateplotcyclelist{anothercustompalette}{
    {EE7F0E!90!black,line width=1.5pt},
    {619D47!90!black,line width=1.5pt},
    {3787CF!90!black,line width=1.5pt},
    {DBDC4A!90!black,line width=1.5pt}
}

\usepackage{stmaryrd}
\usepackage{trimclip}

\makeatletter
\DeclareRobustCommand{\shortto}{%
  \mathrel{\mathpalette\short@to\relax}%
}

\newcommand{\short@to}[2]{%
  \mkern2mu
  \clipbox{{.3\width} 0 0 0}{$\m@th#1\vphantom{+}{\shortrightarrow}$}%
  }
\makeatother

\usepackage[pagebackref=true,breaklinks=true,letterpaper=true,colorlinks,bookmarks=false]{hyperref}

\cvprfinalcopy

\ifcvprfinal\fi

\begin{document}

\makeatletter
\g@addto@macro\@maketitle{
    \vspace*{-21pt}
    \begin{center}\centering
        \setlength{\tabcolsep}{0.06cm}
        \setlength{\itemwidth}{2.81cm}
        \hspace*{-\tabcolsep}\begin{tabular}{cccccc}
                \animategraphics[width=\itemwidth, poster=3, autoplay, palindrome, final, nomouse, method=widget]{12}{graphics/teaser/sepconv/}{00000}{00006}
            &
                \animategraphics[width=\itemwidth, poster=3, autoplay, palindrome, final, nomouse, method=widget]{12}{graphics/teaser/toflow/}{00000}{00006}
            &
                \animategraphics[width=\itemwidth, poster=3, autoplay, palindrome, final, nomouse, method=widget]{12}{graphics/teaser/cyclicgen/}{00000}{00006}
            &
                \animategraphics[width=\itemwidth, poster=3, autoplay, palindrome, final, nomouse, method=widget]{12}{graphics/teaser/ctxsyn/}{00000}{00006}
            &
                \animategraphics[width=\itemwidth, poster=3, autoplay, palindrome, final, nomouse, method=widget]{12}{graphics/teaser/dain/}{00000}{00006}
            &
                \animategraphics[width=\itemwidth, poster=3, autoplay, palindrome, final, nomouse, method=widget]{12}{graphics/teaser/ours/}{00000}{00006}
            \vspace{-0.1cm} \\
                \footnotesize SepConv - $\mathcal{L}_F$~\cite{Niklaus_ICCV_2017}
            &
                \footnotesize ToFlow~\cite{Xue_IJCV_2019}
            &
                \footnotesize CyclicGen~\cite{Liu_AAAI_2019}
            &
                \footnotesize CtxSyn - $\mathcal{L}_F$~\cite{Niklaus_CVPR_2018}
            &
                \footnotesize DAIN~\cite{Bao_CVPR_2019}
            &
                \footnotesize Ours - $\mathcal{L}_F$
            \\
        \end{tabular}\vspace{-0.2cm}
    	\captionof{figure}{A difficult example for video frame interpolation. Our approach produces a high-quality result in spite of the delicate flamingo leg that is subject to large motion. \textit{This is a video figure that is best viewed using Adobe Reader.}}\vspace{-0.0cm}
    	\label{fig:teaser}
    \end{center}
    \vspace*{4pt}
}
\makeatother

\title{Softmax Splatting for Video Frame Interpolation}

\author{Simon Niklaus\\
Portland State University\\
{\tt\small sniklaus@pdx.edu}
\and
Feng Liu\\
Portland State University\\
{\tt\small fliu@cs.pdx.edu}
}
\maketitle
\thispagestyle{empty}

\begin{abstract}

    Differentiable image sampling in the form of backward warping has seen broad adoption in tasks like depth estimation and optical flow prediction. In contrast, how to perform forward warping has seen less attention, partly due to additional challenges such as resolving the conflict of mapping multiple pixels to the same target location in a differentiable way. We propose softmax splatting to address this paradigm shift and show its effectiveness on the application of frame interpolation. Specifically, given two input frames, we forward-warp the frames and their feature pyramid representations based on an optical flow estimate using softmax splatting. In doing so, the softmax splatting seamlessly handles cases where multiple source pixels map to the same target location. We then use a synthesis network to predict the interpolation result from the warped representations. Our softmax splatting allows us to not only interpolate frames at an arbitrary time but also to fine tune the feature pyramid and the optical flow. We show that our synthesis approach, empowered by softmax splatting, achieves new state-of-the-art results for video frame interpolation.

\end{abstract}

\vspace{-0.2cm}\section{Introduction}
\label{sec:intro}
Video frame interpolation is a classic problem in computer vision with many practical applications. It can, for example, be used to convert the frame rate of a video to the refresh rate of the monitor that is used for playback, which is beneficial for human perception~\cite{Kuroki_OTHER_2007, Kuroki_OTHER_2014}. Frame interpolation can also help in video editing tasks, such as temporally consistent color modifications, by propagating the changes that were made in a few keyframes to the remaining frames~\cite{Meyer_BMVC_2018}. Frame interpolation can also support inter-frame compression for videos~\cite{Wu_ECCV_2018}, serve as an auxiliary task for optical flow estimation~\cite{Long_ECCV_2016, Wulff_OTHER_2018}, or generate training data to learn how to synthesize motion blur~\cite{Brooks_CVPR_2019}. While these applications employ frame interpolation in the temporal domain, it can also be used to synthesize novel views in space by interpolating between given viewpoints~\cite{Kalantari_TOG_2016}.

Approaches for video frame interpolation can be categorized as flow-based, kernel-based, and phase-based. We adopt the flow-based paradigm since it has proven to work well in quantitative benchmarks~\cite{Baker_IJCV_2011}. One common approach for these methods is to estimate the optical flow $F_{t \shortto 0}$ and $F_{t \shortto 1}$ between two input frames $I_0$ and $I_1$ from the perspective of the frame $I_t$ that is ought to be synthesized. The interpolation result can then be obtained by backward warping $I_0$ according to $F_{t \shortto 0}$ and $I_1$ according to $F_{t \shortto 1}$~\cite{Jaderberg_NIPS_2015}. While it is intuitive, this approach makes it difficult to use an off-the-shelf optical flow estimator and prevents synthesizing frames at an arbitrary $t$ in a natural manner. To address these concerns, Jiang~\etal~\cite{Jiang_CVPR_2018} and Bao~\etal~\cite{Bao_CVPR_2019} approximate $F_{t \shortto 0}$ and $F_{t \shortto 1}$ from $F_{0 \shortto 1}$ and $F_{1 \shortto 0}$.

Different from backward warping, Niklaus~\etal~\cite{Niklaus_CVPR_2018} directly forward-warp $I_0$ according to $t \cdot F_{0 \shortto 1}$ and $I_1$ according to $(1 - t) \cdot F_{1 \shortto 0}$, which avoids having to approximate $F_{t \shortto 0}$ and $F_{t \shortto 1}$. Another aspect of their approach is to warp not only the images but also the corresponding context information, which a synthesis network can use to make better predictions. However, their forward warping uses the equivalent of z-buffering in order to handle cases where multiple source pixels map to the same target location. It is thus unclear how to fully differentiate this operation due to the z-buffering~\cite{Nguyen_NIPS_2018}. We propose softmax splatting to address this limitation, which allows us to jointly supervise all inputs to the forward warping. As a consequence, we are able to extend the idea of warping a generic context map to learning and warping a task-specific feature pyramid. Furthermore, we are able to supervise not only the optical flow estimator but also the metric that weights the importance of different pixels when they are warped to the same location. This approach, which is enabled by our proposed softmax splatting, achieves new state-of-the-art results and ranks first in the Middlebury benchmark for frame interpolation.

In short, we propose softmax splatting to perform differentiable forward warping and show its effectiveness on the application of frame interpolation. An interesting research question that softmax splatting addresses is how to handle different source pixels that map to the same target location in a differentiable way. Softmax splatting enables us to train and use task-specific feature pyramids for image synthesis. Furthermore, softmax splatting not only allows us to fine-tune an off-the-shelf optical flow estimator for video frame interpolation, it also enables us to supervise the metric that is used to disambiguate cases where multiple source pixels map to the same forward-warped target location.

\section{Related Work}
\label{sec:related}
With the introduction of spatial transformer networks, Jaderberg~\etal~\cite{Jaderberg_NIPS_2015} proposed differentiable image sampling. Since then, this technique has found broad adoption in the form of backward warping to synthesize an image $I_A$ from an image $I_B$ given a correspondence $F_{A \shortto B}$ for each pixel in $I_A$ to its location in $I_B$. Prominent examples where this approach has been used include unsupervised depth estimation~~\cite{Godard_CVPR_2017, Mahjourian_CVPR_2018, Zhou_CVPR_2017}, unsupervised optical flow prediction~\cite{Meister_AAAI_2018, Wang_CVPR_2018, Yu_OTHER_2016}, optical flow prediction~\cite{Hui_CVPR_2018, Ranjan_CVPR_2017, Sun_CVPR_2018}, novel view synthesis~\cite{Cun_OTHER_2019, Liu_CVPR_2018, Zhou_ECCV_2016}, video frame interpolation~\cite{Bao_CVPR_2019, Jiang_CVPR_2018, Liu_AAAI_2019, Liu_ICCV_2017}, and video enhancement~\cite{Caballero_CVPR_2017, Tao_ICCV_2017, Xue_IJCV_2019}.

In contrast, performing forward warping to synthesize $I_B$ from $I_A$ based on $F_{A \shortto B}$ has seen less adoption with deep learning, partly due to additional challenges such as multiple source pixels in $I_A$ possibly being mapped to the same target location in $I_B$. For optical flow estimation, Wang~\etal~\cite{Wang_CVPR_2018} forward-warp an image filled with ones to obtain an occlusion mask. However, they sum up contributions of all the pixels that are mapped to the same output pixel without a mechanism to remove possible outliers, which limits the applicability of this technique for image synthesis. For frame interpolation, Niklaus~\etal~\cite{Niklaus_CVPR_2018} use the equivalent of z-buffering which is well motivated but not differentiable~\cite{Nguyen_NIPS_2018}. Bao~\etal~\cite{Bao_CVPR_2019} linearly weight the optical flow according to a depth estimate as an approach for dealing with multiple source pixels mapping to the same target location. However, adding a bias to the depth estimation affects the result of this linearly weighted warping and leads to negative side effects. In contrast, our proposed softmax splatting is not subject to any of these concerns.

We demonstrate the effectiveness of our proposed softmax splatting on the example of frame interpolation. Research on frame interpolation has seen a recent resurgence, with multiple papers proposing kernel-based~\cite{Bao_CVPR_2019, Bao_ARXIV_2018, Niklaus_CVPR_2017, Niklaus_ICCV_2017}, flow-based~\cite{Bao_CVPR_2019, Bao_ARXIV_2018, Jiang_CVPR_2018, Liu_AAAI_2019, Liu_ICCV_2017, Niklaus_CVPR_2018, Raket_OTHER_2012, Reda_ICCV_2019, Xue_IJCV_2019}, and phase-based~\cite{Meyer_CVPR_2018, Meyer_CVPR_2015} approaches. We base our approach on the one from Niklaus~\etal~\cite{Niklaus_CVPR_2018} who estimate optical flow between two input images in both directions, extract generic contextual information from the input images using pre-trained filters, forward-warp the images together with their context maps according to optical flow, and finally employ a synthesis network to obtain the interpolation result. Enabled by softmax splatting, we extend their framework to warping task-specific feature pyramids for image synthesis in an end-to-end manner. This includes fine-tuning the off-the-shelf optical flow estimator for video frame interpolation and supervising the metric that is used to disambiguate cases where multiple pixels map to the same location.

\begin{figure}\centering
    \setlength{\tabcolsep}{0.2cm}
    \setlength{\itemwidth}{4.25cm}
    \hspace*{-\tabcolsep}\begin{tabular}{cc}
            \includegraphics[]{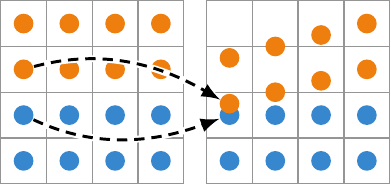}
        &
            \includegraphics[]{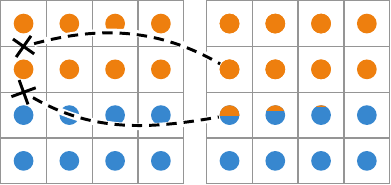}
        \\
            \footnotesize forward warping / splatting
        &
            \footnotesize backward warping / sampling
        \\
    \end{tabular}\vspace{-0.2cm}
    \caption{Splatting versus sampling, the blue pixels remain static while the red ones move down in a shearing manner. With splatting, the output is subject to holes and multiple source pixels can map to the same target pixel. On the upside, splatting makes it possible to scale the transform.}\vspace{-0.3cm}
	\label{fig:wapring}
\end{figure}

\begin{figure*}\centering
    \setlength{\tabcolsep}{0.08cm}
    \setlength{\itemwidth}{4.25cm}
    \hspace*{-\tabcolsep}\begin{tabular}{cccc}
            \animategraphics[width=\itemwidth, poster=10, autoplay, palindrome, final, nomouse, method=widget]{12}{graphics/warping/summa/}{00000}{00010}
        &
            \animategraphics[width=\itemwidth, poster=10, autoplay, palindrome, final, nomouse, method=widget]{12}{graphics/warping/averg/}{00000}{00010}
        &
            \animategraphics[width=\itemwidth, poster=10, autoplay, palindrome, final, nomouse, method=widget]{12}{graphics/warping/linea/}{00000}{00010}
        &
            \animategraphics[width=\itemwidth, poster=10, autoplay, palindrome, final, nomouse, method=widget]{12}{graphics/warping/propo/}{00000}{00010}
        \vspace{-0.1cm} \\
            \footnotesize summation splatting $\overrightarrow{\Sigma}$
        &
            \footnotesize average splatting $\overrightarrow{\Phi}$
        &
            \footnotesize linear splatting $\overrightarrow{\ast}$
        &
            \footnotesize softmax splatting $\overrightarrow{\sigma}$
        \\
    \end{tabular}\vspace{-0.2cm}
    \caption{Given two images $I_0$ and $I_1$ as well as an optical flow estimate $F_{0 \shortto 1}$, this figure shows an example of warping $I_0$ to $I_t$ according to $F_{0 \shortto t} = t \cdot F_{0 \shortto 1}$ with four different forward warping approaches. The summation warping $\protect\overrightarrow{\Sigma}$ handles cases where multiple pixels in $I_0$ map to the same target location in $I_t$ by taking their sum, which leads to brightness inconsistencies. The average warping $\protect\overrightarrow{\Phi}$ takes their mean instead and is able to maintain the overall appearance of $I_0$ but blends overlapping regions. The linear splatting $\protect\overrightarrow{\ast}$ weights the pixels in $I_0$ before warping them but still fails to clearly separate the front of the car from the grass in the background. In contrast, our proposed softmax splatting $\protect\overrightarrow{\sigma}$ shows the expected behavior with the car correctly occluding the background. \textit{This is a video figure that is best viewed using Adobe Reader.}}\vspace{-0.3cm}
	\label{fig:splatting}
\end{figure*}

For image synthesis, Niklaus~\etal~\cite{Niklaus_CVPR_2018} warp context information from a pre-trained feature extractor that a synthesis network can use to make better predictions. Bao~\etal~\cite{Bao_CVPR_2019} subsequently refined this approach through end-to-end supervision of the feature extractor. In contrast, we extract and warp feature pyramids which allows the synthesis network to make use of a multi-scale representation for better interpolation results. Our use of feature pyramids for image synthesis is inspired by recent work on video analysis. For video semantic segmentation, Gadde~\etal~\cite{Gadde_ICCV_2017} warp features that were obtained when processing the preceding frame in order to support the segmentation of the current frame. For optical flow estimation, Hui~\etal~\cite{Hui_CVPR_2018} and Sun~\etal~\cite{Sun_CVPR_2018} extend this idea of warping features and employ it across multiple scales in the form of feature pyramids. These approaches do not target image synthesis though.

Temporal consistency is a common concern when synthesizing images in time~\cite{Aydin_TOG_2014, Huang_CVPR_2017, Huang_TOG_2016, Lai_ECCV_2018}. For frame interpolation, Jiang~\etal~\cite{Jiang_CVPR_2018} collect a specialized training dataset with frame-nonuples and supervise their network on seven intermediate frames at a time in order to ensure temporally consistent results. In the same vein, Liu~\etal~\cite{Liu_AAAI_2019} and Reda~\etal~\cite{Reda_ICCV_2019} utilize cycle consistency to better supervise their model. In comparison, our proposed softmax splatting leads to temporally consistent results without requiring a specialized training dataset or cycle-consistent training.

\section{Softmax Splatting for Frame Interpolation}
\label{sec:method}
Given two frames $I_0$ and $I_1$, frame interpolation aims to synthesize an intermediate frame $I_t$ where $t \in \left( 0, 1 \right)$ defines the desired temporal position. To address this problem, we first use an off-the-shelf optical flow method to estimate the optical flow $F_{0 \shortto 1}$ and $F_{1 \shortto 0}$ between the input frames in both directions. We then use forward warping in the form of softmax splatting $\overrightarrow{\sigma}$ to warp $I_0$ according to $F_{0 \shortto t} = t \cdot F_{0 \shortto 1}$ and $I_1$ according to $F_{1 \shortto t} = (1 - t) \cdot F_{1 \shortto 0}$ as follows.
\begin{align} \begin{split}
    I_t & \approx \overrightarrow{\sigma} \left( I_0, F_{0 \shortto t} \right) = \overrightarrow{\sigma} \left( I_0, t \cdot F_{0 \shortto 1} \right)
\end{split} \\ \begin{split}
    I_t & \approx \overrightarrow{\sigma} \left( I_1, F_{1 \shortto t} \right) = \overrightarrow{\sigma} \left( I_1, (1 - t) \cdot F_{1 \shortto 0} \right)
\end{split} \end{align}
This is in contrast to backward warping $\overleftarrow{\omega}$, which would require $F_{t \shortto 0}$ and $F_{t \shortto 1}$ but computing this $t$-centric optical flow from $F_{0 \shortto 1}$ and $F_{1 \shortto 0}$ is complicated and subject to approximations~\cite{Bao_CVPR_2019}. We then combine these intermediate results to obtain $I_t$ using a synthesis network. More specifically, we not only warp the input frame in color- but also feature-space across multiple resolutions which enables the synthesis network to make better predictions.

We subsequently first introduce forward warping via softmax splatting and then show how it enables us to establish new state-of-the-art results for frame interpolation.

\subsection{Forward Warping via Softmax Splatting}\label{sec:splatting}

Backward warping is a common technique that has found broad adoption in tasks like unsupervised depth estimation or optical flow prediction~\cite{Jaderberg_NIPS_2015}. It is well supported by many deep learning frameworks. In contrast, forward warping an image $I_0$ to $I_t$ according to $F_{0 \shortto t}$ is not supported by these frameworks. We attribute this lack of support to the fact that there is no definitive way of performing forward warping. Forward warping is subject to multiple pixels in $I_0$ being able to possibly map to the same target pixel in $I_t$ and there are various possibilities to address this ambiguity. We thus subsequently introduce common approaches to handle this mapping-ambiguity and discuss their limitations. We then propose softmax splatting which addresses these inherent limitations. Please note that we use the terms ``forward warping'' and ``splatting'' interchangeably.

\vspace{0.05in}
\noindent\textbf{Summation splatting.} A straightforward approach of handling the aforementioned mapping-ambiguity is to sum all contributions. We define this summation splatting $\overrightarrow{\Sigma}$ as follows, where $I_t^\Sigma$ is the sum of all contributions from $I_0$ to $I_t$ according to $F_{0 \shortto t}$ subject to the bilinear kernel $b$.
\begin{align} \begin{split}
    \text{let } \bm{u} & = \bm{p} - \big( \bm{q} + F_{0 \shortto t}[\bm{q}] \big)
\end{split} \\[0.09cm] \begin{split}
    b(\bm{u}) & = \text{max} \mleft( 0, 1 - \left| \bm{u}_x \right| \mright) \cdot \text{max} \mleft( 0, 1 - \left| \bm{u}_y \right| \mright)
\end{split} \\[0.09cm] \begin{split}
    I_t^\Sigma[\bm{p}] & = \sum_{\forall \bm{q} \in I_0} b(\bm{u}) \cdot I_0[\bm{q}]
\end{split} \\[0.09cm] \begin{split}
    \overrightarrow{\Sigma} & \left( I_0,  F_{0 \shortto t} \right)= I_t^\Sigma
\end{split} \end{align}
As shown in Figure~\ref{fig:splatting}, this summation splatting leads to brightness inconsistencies in overlapping regions like the front of the car. Furthermore, the bilinear kernel $b$ leads to pixels in $I_t$ that only receive partial contributions from the pixels in $I_0$ which yet again leads to brightness inconsistencies like on the street. However, we use this summation splatting as the basis of all subsequent forward warping approaches. The relevant derivatives are as follows.
\begin{align} \begin{split}
    \text{let } \bm{u} & = \bm{p} - \big( \bm{q} + F_{0 \shortto t}[\bm{q}] \big)
\end{split} \\[0.09cm] \begin{split}
    \frac{\partial I_t^\Sigma[\bm{p}]}{\partial I_0[\bm{q}]} & = b(\bm{u})
\end{split} \\ \begin{split}
    \hspace{-0.1cm} \frac{\partial I_t^\Sigma[\bm{p}]}{\partial F_{0 \shortto t}^x[\bm{q}]} & = \frac{\partial b(\bm{u})}{\partial F_{0 \shortto t}^x} \cdot I_0[\bm{q}]
\end{split} \\ \begin{split}
    \frac{\partial b(\bm{u})}{\partial F_{0 \shortto t}^x} & = \text{max}\mleft( 0, 1 - \left| \bm{u}_y \right| \mright) \cdot \begin{cases}
        0, \hspace{0.1cm} \text{if } \left| \bm{u}_x \right| \geq 1
        \\
        -\text{sgn}(\bm{u}_x), \hspace{0.1cm} \text{else}
    \end{cases}
\end{split} \end{align}
and analogous for the \emph{y} component of $F_{0 \shortto t}$. It is not easy to obtain these through automatic differentiation since few frameworks support the underlying \texttt{scatter\_nd} function that is necessary to implement this operator. We hence provide a PyTorch reference implementation\footnote{\url{http://sniklaus.com/softsplat}} of this summation splatting $\overrightarrow{\Sigma}$ which is written in CUDA for efficiency.

\vspace{0.05in}
\noindent\textbf{Average splatting.} To address the brightness inconsistencies that occur with summation splatting, we need to normalize $I_t^\Sigma$. To do so, we can reuse the definition of $\overrightarrow{\Sigma}$ and determine average splatting $\overrightarrow{\Phi}$ as follows.
\begin{equation}\begin{aligned}
    \overrightarrow{\Phi} \left( I_0, F_{0 \shortto t} \right) = \frac{ \overrightarrow{\Sigma} \left( I_0, F_{0 \shortto t} \right) }{ \overrightarrow{\Sigma} \left( \bm{1}, F_{0 \shortto t} \right) }
\end{aligned}\end{equation}
As shown in Figure~\ref{fig:splatting}, this approach handles the brightness inconsistencies and maintains the appearance of $I_0$. However, this technique averages overlapping regions like at the front of the car with the grass in the background.

\vspace{0.05in}
\noindent\textbf{Linear splatting.} In an effort to better separate overlapping regions, one could try to linearly weight $I_0$ by an importance mask $Z$ and define linear splatting $\overrightarrow{\ast}$ as follows.
\begin{equation}\begin{aligned}
    \overrightarrow{\ast} \left( I_0, F_{0 \shortto t} \right) = \frac{ \overrightarrow{\Sigma} \left( Z \cdot I_0, F_{0 \shortto t} \right) }{ \overrightarrow{\Sigma} \left( Z, F_{0 \shortto t} \right) }
\end{aligned}\end{equation}
where $Z$ could, for example, relate to the depth of each pixel~\cite{Bao_CVPR_2019}. As shown in Figure~\ref{fig:splatting}, this approach can better separate the front of the car from the grass in the background. It is not invariant to translations with respect to $Z$ though. If $Z$ represents the inverse depth then there will be a clear separation if the car is at $Z = \nicefrac{1}{1}$ and the background is at $Z = \nicefrac{1}{10}$. But, if the car is at $Z = \nicefrac{1}{101}$ and the background is at $Z = \nicefrac{1}{110}$ then they will be averaged again despite being equally far apart in terms of depth.

% \vspace{0.05in}
\noindent\textbf{Softmax splatting.} To clearly separate overlapping regions according to an importance mask $Z$ with translational invariance, we propose softmax splatting $\overrightarrow{\sigma}$ as follows.
\begin{equation}\begin{aligned}
    \overrightarrow{\sigma} \left( I_0, F_{0 \shortto t} \right) = \frac{ \overrightarrow{\Sigma} \left( \exp(Z) \cdot I_0, F_{0 \shortto t} \right) }{ \overrightarrow{\Sigma} \left( \exp(Z), F_{0 \shortto t} \right) }
\end{aligned}\end{equation}
where Z could, for example, relate to the depth of each pixel~\cite{Bao_CVPR_2019}. As shown in Figure~\ref{fig:splatting}, this approach is able to clearly separate the front of the car from the background without any remaining traces of grass. Furthermore, it shares resemblance to the softmax function. It is hence invariant to translations $\beta$ with respect to $Z$, which is a particularly important property when mapping multiple pixels to the same location. If $Z$ represents depth, then the car and the background in Figure~\ref{fig:splatting} are treated equally whether the car is at $Z = 1$ and the background is at $Z = 10$ or the car is at $Z = 101$ and the background is at $Z = 110$. It is not invariant to scale though and multiplying $Z$ by $\alpha$ will affect how well overlapping regions will be separated. A small $\alpha$ yields averaging whereas a large $\alpha$ yields z-buffering. This parameter can be learned via end-to-end training.

\vspace{0.05in}
\noindent\textbf{Importance metric.} We use $Z$ to weight pixels in $I_0$ in order to resolve cases where multiple pixels from $I_0$ map to the same target pixel in $I_t$. This $Z$ could, for example, represent depth~\cite{Bao_CVPR_2019}. However, obtaining such a depth estimate is computationally expensive and inherently challenging which makes it prone to inaccuracies. We thus use brightness constancy as a measure of occlusion~\cite{Baker_IJCV_2011}, which can be obtained via backward warping $\overleftarrow{\omega}$ as follows.
\begin{equation}\begin{aligned}
    Z = \alpha \cdot \left\| I_0 - \overleftarrow{\omega} \left( I_1, F_{0 \shortto 1} \right) \right\| _1
\end{aligned}\end{equation}
Since our proposed softmax splatting is fully differentiable, we can not only learn $\alpha$ (initially set to $-1$) but also use a small neural network $\upsilon$ to further refine this metric.
\begin{equation}\begin{aligned}
    Z = \upsilon \big( I_0, - \left\| I_0 - \overleftarrow{\omega} \left( I_1, F_{0 \shortto 1} \right) \right\| _1 \big)
\end{aligned}\end{equation}
One could also obtain $Z$ directly from $\upsilon ( I_0 )$ but we were unable to make this $\upsilon$ converge. Lastly, when applying softmax splatting to tasks different from frame interpolation, the importance metric may be adjusted accordingly.

\vspace{0.05in}
\noindent\textbf{Efficiency.} PyTorch's backward warping requires $1.1$ ms to warp a full-HD image on a Titan X with a synthetic flow drawn from $\mathcal{N}(0, 10^2)$. In contrast, our implementation of softmax splatting requires $3.7$ ms since we need to compute $Z$ and handle race conditions during warping.

\begin{figure*}\centering
    \vspace{-0.1cm}\includegraphics[]{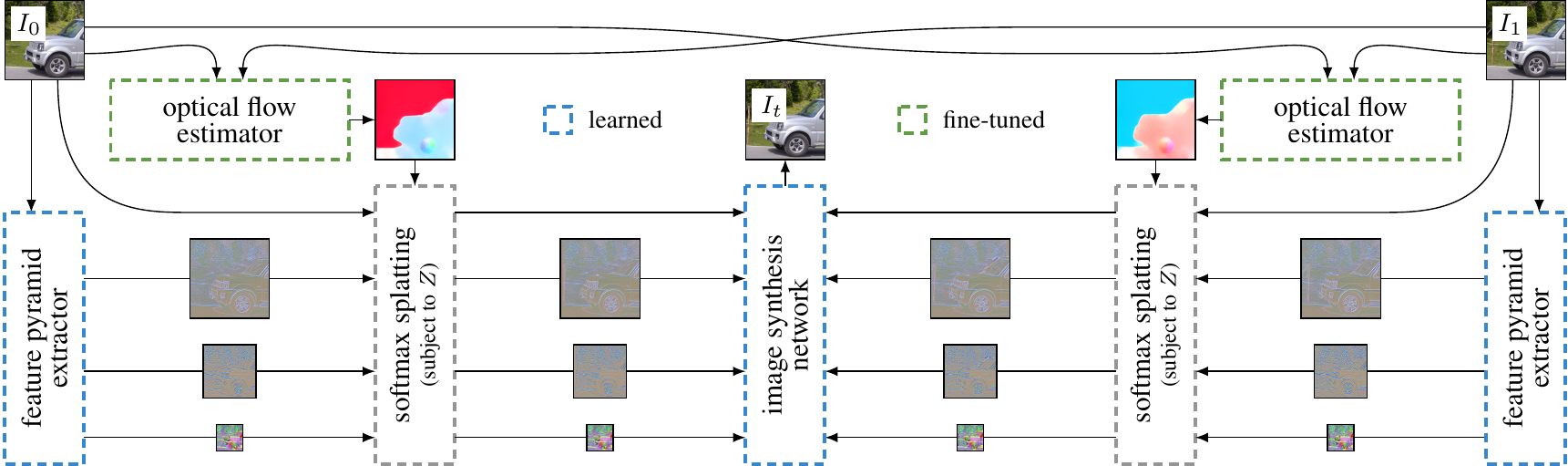}\vspace{-0.1cm}
	\caption{An overview of our frame interpolation framework. Given two input frames $I_0$ and $I_1$, we first estimate the bidirectional optical flow between them. We then extract their feature pyramids and forward-warp them together with the input frames to the target temporal position $t \in \left( 0, 1 \right)$ according to the optical flow. Using softmax splatting enables end-to-end training and thus allows the feature pyramid extractor to learn to gather features that are important for image synthesis. The warped input frames and feature pyramids are then fed to a synthesis network to generate the interpolation result $I_t$.}\vspace{-0.3cm}
	\label{fig:architecture}
\end{figure*}

\subsection{Feature Pyramids for Image Synthesis}

We adopt the video frame interpolation pipeline from Niklaus~\etal~\cite{Niklaus_CVPR_2018} who, given two input frames $I_0$ and $I_1$, first estimate the inter-frame motion $F_{0 \shortto 1}$ and $F_{1 \shortto 0}$ using an off-the-shelf optical flow method. They then extract generic contextual information from the input images using a pre-defined filter $\psi$ and forward-warp $\overrightarrow{\omega}$ the images together with their context maps according to $t \cdot F_{0 \shortto 1} = F_{0 \shortto t}$ and $(1 - t) \cdot F_{1 \shortto 0} = F_{1 \shortto t}$, before employing a synthesis network $\phi$ to obtain the interpolation result $I_t$.
\begin{equation*}
    I_t = \phi \Big( \overrightarrow{\omega} \big( \{ I_0, \psi \left( I_0 \right) \}, F_{0 \shortto t} \big), \overrightarrow{\omega} \big( \{ I_1, \psi \left( I_1 \right) \}, F_{1 \shortto t} \big) \Big)
\end{equation*}
This approach is conceptually simple and has been proven to work well. However, Niklaus~\etal were not able to supervise the context extractor $\psi$ and instead used \texttt{conv1} of ResNet-18~\cite{He_CVPR_2016} due to the limitations of their forward warping $\overrightarrow{\omega}$ approach. This limitation makes it an ideal candidate to show the benefits of our proposed softmax splatting.

Our proposed softmax splatting allows us to supervise $\psi$, enabling it to learn to extract features that are important for image synthesis. Furthermore, we extend this idea by extracting and warping features at multiple scales in the form of feature pyramids. This allows the synthesis network $\phi$ to further improve its predictions. Please see Figure~\ref{fig:architecture} for an overview of our video frame interpolation framework. We will subsequently discuss its individual components.

\begin{figure}\centering
    \includegraphics[]{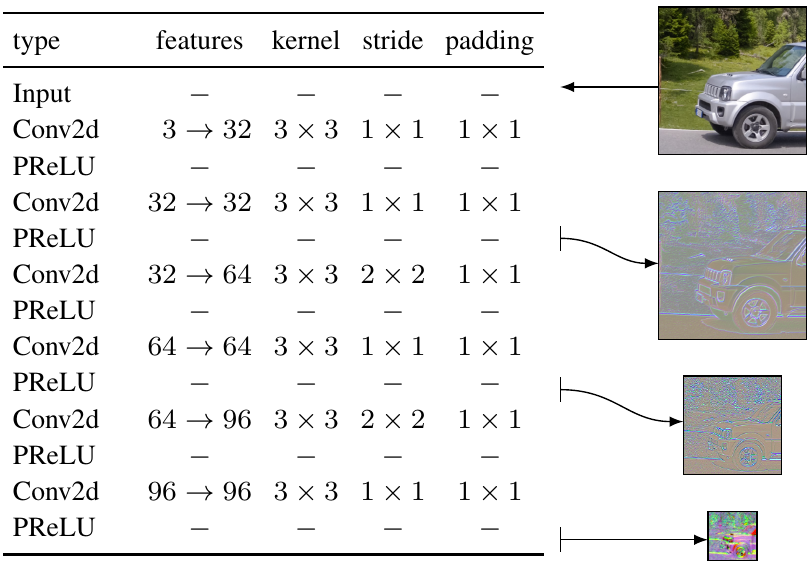}\vspace{-0.1cm}
	\caption{The architecture of our feature pyramid extractor. The feature visualization was obtained using PCA and is only serving an aesthetic purpose. See our evaluation for an analysis of the feature pyramid space for image synthesis.}\vspace{-0.3cm}
	\label{fig:pyramid}
\end{figure}

\vspace{0.05in}
\noindent\textbf{Optical flow estimator.} We use an off-the-shelf optical flow method to make use of the ongoing achievements in research on correspondence estimation. Specifically, we use PWC-Net~\cite{Sun_CVPR_2018} and show that FlowNet2~\cite{Ilg_CVPR_2017} and LiteFlowNet~\cite{Hui_CVPR_2018} perform equally well within our evaluation. In accordance with the findings of Xue~\etal~\cite{Xue_IJCV_2019}, we additionally fine-tune PWC-Net for frame interpolation.

\vspace{0.05in}
\noindent\textbf{Feature pyramid extractor.} The architecture of our feature pyramid extractor is shown in Figure~\ref{fig:pyramid}. Our proposed softmax splatting enables us to supervise this feature pyramid extractor in an end-to-end manner, allowing it to learn to extract features that are useful for the subsequent image synthesis. As shown in our evaluation, this approach leads to significant improvements in the quality of the interpolation result. We also show that the interpolation quality degrades if we use fewer levels of features.

\vspace{0.05in}
\noindent\textbf{Image synthesis network.} The synthesis network generates the interpolation result guided by the warped input images and their corresponding feature pyramids. We employ a GridNet~\cite{Fourure_BMVC_2017} architecture with three rows and six columns for this task. To avoid checkerboard artifacts~\cite{Odena_OTHER_2016}, we adopt the modifications proposed by Niklaus~\etal~\cite{Niklaus_CVPR_2018}. The GridNet architecture is a generalization of U-Nets and is thus well suited for the task of image synthesis.

\vspace{0.05in}
\noindent\textbf{Importance metric.} Our proposed softmax splatting uses an importance metric $Z$ which is used to resolve cases where multiple pixels forward-warp to the same target location. We use brightness constancy to compute this metric as outlined in Section~\ref{sec:splatting}. Furthermore, we refine this occlusion estimate using a small U-Net consisting of three levels, which is trained end-to-end with the feature pyramid extractor and the image synthesis network.

\vspace{0.05in}
\noindent\textbf{Training.} We adopt the training from Niklaus~\etal~\cite{Niklaus_CVPR_2018}. We thus train two versions of our model to account for the perception-distortion tradeoff~\cite{Blau_CVPR_2018}, one trained on color loss $\mathcal{L}_{\textit{Lap}}$ which performs well in standard benchmarks and one trained on perceptual loss $\mathcal{L}_F$ which retains more details in difficult cases. However, instead of using a proprietary training dataset, we use frame-triples from the training portion of the publicly available Vimeo-90k dataset~\cite{Xue_IJCV_2019}.

\vspace{0.05in}
\noindent\textbf{Efficiency.} With an Nvidia Titan X, we are able to synthesize a 720p frame in $0.357$ seconds as well as a 1080p frame in $0.807$ seconds. The parameters of our entire pipeline amount to $31$ megabytes when stored.

\section{Experiments}
\label{sec:experiments}
We evaluate our method, which utilizes softmax splatting to improve an existing frame interpolation approach, and compare it to state-of-the-art methods quantitatively and qualitatively on publicly available datasets. To support examining the visual quality of the frame interpolation results, we additionally provide a supplementary video.

\vspace{0.05in}
\noindent\textbf{Methods.} We compare our approach to several state-of-the-art frame interpolation methods for which open source implementations from the respective authors are publicly available. This includes SepConv~\cite{Niklaus_ICCV_2017}, ToFlow~\cite{Xue_IJCV_2019}, CyclicGen~\cite{Liu_AAAI_2019}, and DAIN~\cite{Bao_CVPR_2019}. We also include the closed source CtxSyn~\cite{Niklaus_CVPR_2018} approach wherever possible.

\vspace{0.05in}
\noindent\textbf{Datasets.} We perform the quantitative evaluation on common datasets for frame interpolation. This includes the Vimeo-90k~\cite{Xue_IJCV_2019} test dataset as well as the samples from the Middlebury benchmark with publicly-available ground truth interpolation results~\cite{Baker_IJCV_2011}. When comparing our approach to other state-of-the-art methods, we additionally incorporate samples from UCF101~\cite{Liu_ICCV_2017, Soomro_ARXIV_2012} and Xiph\footnote{\url{https://media.xiph.org/video/derf}}.

\vspace{0.05in}
\noindent\textbf{Metrics.} We follow recent work on frame interpolation and use PSNR and SSIM~\cite{Wang_TIP_2004} for all quantitative comparisons. We additionally incorporate the LPIPS~\cite{Rizhang_CVPR_2018} metric which strives to measure perceptual similarity. While higher values indicate better results in terms of PSNR and SSIM, lower values indicate better results with the LPIPS metric.

\begin{figure}\centering
	\setlength{\tabcolsep}{0.0cm}
	\renewcommand{\arraystretch}{1.2}
	\newcommand{\quantTit}[1]{\multicolumn{3}{c}{\scriptsize #1}}
	\newcommand{\quantSec}[1]{\scriptsize #1}
	\newcommand{\quantInd}[1]{\scriptsize #1}
	\newcommand{\quantVal}[1]{\scalebox{0.83}[1.0]{$ #1 $}}
	\newcommand{\quantBes}[1]{\scalebox{0.83}[1.0]{$\uline{ #1 }$}}
	\footnotesize
	\begin{tabularx}{\columnwidth}{@{\hspace{0.1cm}} X P{0.95cm} @{\hspace{-0.2cm}} P{0.95cm} @{\hspace{-0.2cm}} P{0.95cm} P{0.95cm} @{\hspace{-0.2cm}} P{0.95cm} @{\hspace{-0.2cm}} P{0.95cm}}
		\toprule
			& \quantTit{Vimeo-90k~\cite{Xue_IJCV_2019}} & \quantTit{Middlebury~\cite{Baker_IJCV_2011}}
		\\ \cmidrule(l{2pt}r{2pt}){2-4} \cmidrule(l{2pt}r{2pt}){5-7}
			& \quantSec{PSNR} \linebreak \quantInd{$\uparrow$} & \quantSec{SSIM} \linebreak \quantInd{$\uparrow$} & \quantSec{LPIPS} \linebreak \quantInd{$\downarrow$} & \quantSec{PSNR} \linebreak \quantInd{$\uparrow$} & \quantSec{SSIM} \linebreak \quantInd{$\uparrow$} & \quantSec{LPIPS} \linebreak \quantInd{$\downarrow$}
		\\ \midrule
CtxSyn & \quantVal{34.39} & \quantVal{0.961} & \quantBes{0.024} & \quantVal{36.93} & \quantVal{0.964} & \quantBes{0.016}
\\
Ours - CtxSyn-like & \quantBes{34.85} & \quantBes{0.963} & \quantVal{0.025} & \quantBes{37.02} & \quantBes{0.966} & \quantVal{0.018}
\\ \midrule
Ours - summation splatting & \quantVal{35.09} & \quantVal{0.965} & \quantVal{0.024} & \quantVal{37.47} & \quantVal{0.968} & \quantVal{0.018}
\\
Ours - average splatting & \quantVal{35.29} & \quantVal{0.966} & \quantBes{0.023} & \quantVal{37.53} & \quantBes{0.969} & \quantBes{0.017}
\\
Ours - linear splatting & \quantVal{35.26} & \quantVal{0.966} & \quantVal{0.024} & \quantVal{37.73} & \quantVal{0.968} & \quantBes{0.017}
\\
Ours - softmax splatting & \quantBes{35.54} & \quantBes{0.967} & \quantVal{0.024} & \quantBes{37.81} & \quantBes{0.969} & \quantBes{0.017}
\\ \midrule
Ours - pre-defined $Z$ & \quantVal{35.54} & \quantBes{0.967} & \quantBes{0.024} & \quantVal{37.81} & \quantVal{0.969} & \quantBes{0.017}
\\
Ours - fine-tuned $Z$ & \quantBes{35.59} & \quantBes{0.967} & \quantBes{0.024} & \quantBes{37.97} & \quantBes{0.970} & \quantBes{0.017}
\\ \midrule
Ours - 1 feature level & \quantVal{35.08} & \quantVal{0.965} & \quantVal{0.024} & \quantVal{37.32} & \quantVal{0.968} & \quantVal{0.018}
\\
Ours - 2 feature levels & \quantVal{35.37} & \quantVal{0.966} & \quantVal{0.024} & \quantVal{37.79} & \quantVal{0.970} & \quantBes{0.016}
\\
Ours - 3 feature levels & \quantVal{35.59} & \quantVal{0.967} & \quantVal{0.024} & \quantVal{37.97} & \quantVal{0.970} & \quantVal{0.017}
\\
Ours - 4 feature levels & \quantBes{35.69} & \quantBes{0.968} & \quantBes{0.023} & \quantBes{37.99} & \quantBes{0.971} & \quantBes{0.016}
\\ \midrule
Ours - FlowNet2 & \quantVal{35.83} & \quantVal{0.969} & \quantVal{0.022} & \quantVal{37.67} & \quantVal{0.970} & \quantBes{0.016}
\\
Ours - LiteFlowNet & \quantVal{35.59} & \quantVal{0.968} & \quantVal{0.024} & \quantVal{37.83} & \quantVal{0.970} & \quantVal{0.017}
\\
Ours - PWC-Net & \quantVal{35.59} & \quantVal{0.967} & \quantVal{0.024} & \quantVal{37.97} & \quantVal{0.970} & \quantVal{0.017}
\\
Ours - PWC-Net-ft & \quantBes{36.10} & \quantBes{0.970} & \quantBes{0.021} & \quantBes{38.42} & \quantBes{0.971} & \quantBes{0.016}
\\ \midrule
Ours - $\mathcal{L}_{\textit{Lap}}$ & \quantBes{36.10} & \quantBes{0.970} & \quantVal{0.021} & \quantBes{38.42} & \quantBes{0.971} & \quantVal{0.016}
\\
Ours - $\mathcal{L}_{\textit{F}}$ & \quantVal{35.48} & \quantVal{0.964} & \quantBes{0.013} & \quantVal{37.55} & \quantVal{0.965} & \quantBes{0.008}
		\\ \bottomrule
	\end{tabularx}\vspace{-0.1cm}
	\captionof{table}{Ablation experiments to quantitatively analyze the effect of the different components of our approach.}\vspace{-0.3cm}
	\label{tbl:ablation}
\end{figure}

\subsection{Ablation Experiments}\label{sec:ablation}

We show the effectiveness of our proposed softmax splatting by improving the context-aware frame interpolation from Niklaus~\etal~\cite{Niklaus_CVPR_2018}. We thus not only need to compare softmax splatting to alternative ways of performing differentiable forward warping, we also need to analyze the improvements that softmax splatting enabled.

\vspace{0.05in}
\noindent\textbf{Context-aware synthesis.} Since we adopt the framework of Niklaus~\etal~\cite{Niklaus_CVPR_2018}, we first need to verify that we can match their performance. We thus replace our feature pyramid extractor with the \texttt{conv1} layer of ResNet-18~\cite{He_CVPR_2016} and we do not fine-tune the utilized PWC-Net for frame interpolation. This leaves the training dataset as well as the softmax splatting as the only significant differences. As shown in Table~\ref{tbl:ablation} (first section), our implementation performs slightly better in terms of PSNR on the Middlebury examples. It is significantly better in terms of PSNR on the Vimeo-90k test data though, but this is to be expected since we supervise on the Vimeo-90k training data. We can thus confirm that the basis for our approach truthfully replicates CtxSyn.

\vspace{0.05in}
\noindent\textbf{Softmax splatting for frame interpolation.} We discussed various ways of performing differentiable forward warping in Section~\ref{sec:splatting} and outlined their limitations. We then proposed softmax splatting to address these limitations. To analyze the effectiveness of softmax splatting, we train four versions of our approach, each one using a different forward warping technique. As shown in Table~\ref{tbl:ablation} (second section), summation splatting performs worst and softmax splatting performs best in terms of PSNR. Notice that the PSNR of average splatting is better than linear splatting on the Middlebury examples but worse on the Vimeo-90k test data. We attribute this erratic behavior of linear splatting to its lack of translational invariance. These findings support the motivations behind our proposed softmax splatting.

\begin{figure}\centering
    \setlength{\tabcolsep}{0.06cm}
    \setlength{\itemwidth}{2.7cm}
    \hspace*{-\tabcolsep}\begin{tabular}{ccc}
            \includegraphics[width=\itemwidth]{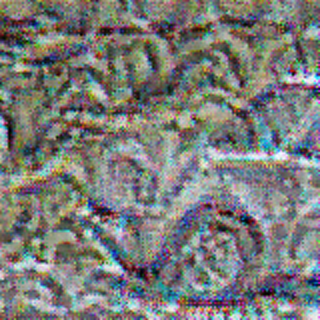}
        &
            \includegraphics[width=\itemwidth]{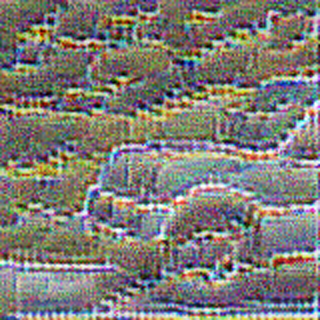}
        &
            \includegraphics[width=\itemwidth]{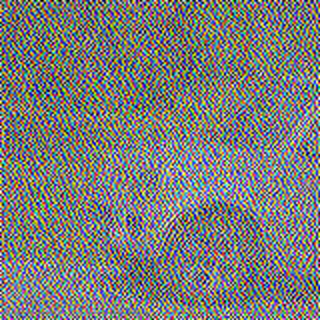}
        \vspace{-0.0cm} \\
            \includegraphics[width=\itemwidth]{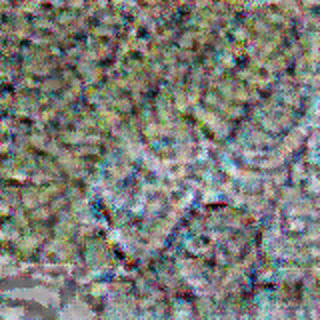}
        &
            \includegraphics[width=\itemwidth]{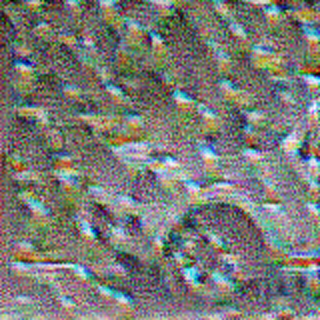}
        &
            \includegraphics[width=\itemwidth]{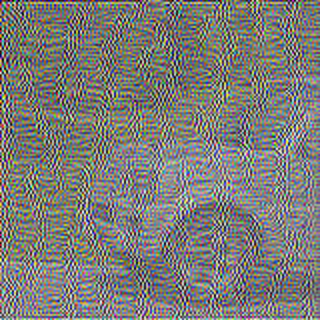}
        \vspace{-0.0cm} \\
            \footnotesize PWC-Net
        &
            \footnotesize LiteFlowNet
        &
            \footnotesize Ours
        \\
    \end{tabular}\vspace{-0.2cm}
	\caption{Feature response visualization for different task-specific feature pyramids on the image from Figure~\ref{fig:splatting} using the visualization technique from Erhan~\etal~\cite{Erhan_OTHER_2009}.}\vspace{-0.3cm}
	\label{fig:features}
\end{figure}

\vspace{0.05in}
\noindent\textbf{Importance metric.} Our proposed softmax splatting uses an importance metric $Z$ to resolve cases where multiple pixels forward-warp to the same target location. We use brightness constancy~\cite{Baker_IJCV_2011} to obtain this metric. Since softmax splatting is fully differentiable, we can use a small U-Net to fine-tune this metric which, as shown in Table~\ref{tbl:ablation} (third section), leads to slight improvements in terms of PSNR. This demonstrates that softmax splatting can effectively supervise $Z$ and that brightness constancy works well as the importance metric for video frame interpolation.

\vspace{0.05in}
\noindent\textbf{Feature pyramids for image synthesis.} Softmax splatting enables us to synthesize images from warped feature pyramids, effectively extending the interpolation framework from Niklaus~\etal~\cite{Niklaus_CVPR_2018}. In doing so, the softmax splatting enables end-to-end training of the feature pyramid extractor, allowing it to learn to gather features that are important for image synthesis. As shown in Table~\ref{tbl:ablation} (fourth section), the quality of the interpolation results improves when using more feature levels. Notice the diminishing returns when using more feature levels, with four levels of features overfitting on the Vimeo-90k dataset. We thus use three levels of features for our approach. We examine the difference between feature pyramids for frame interpolation and those for motion estimation by visualizing their feature responses~\cite{Erhan_OTHER_2009}. Specifically, we maximize the activations of the last layer of our feature pyramid extractor as well as equivalent layers of PWC-Net~\cite{Sun_CVPR_2018} and LiteFlowNet~\cite{Hui_CVPR_2018} by altering the input image. Figure~\ref{fig:features} shows representative feature activations, indicating that our feature pyramid focuses on fine details which are important to synthesize high-quality results while the feature pyramids for optical flow exhibit large patterns to account for large displacements.

\begin{figure}\centering
    \hspace{-0.1cm}\includegraphics[]{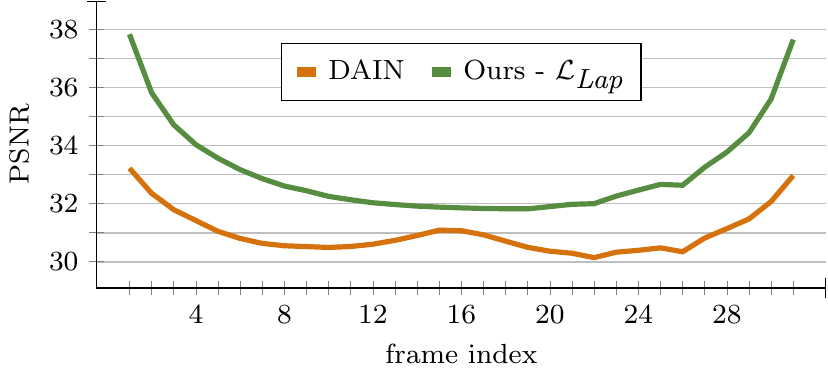}\vspace{-0.2cm}
	\caption{Assessment of the temporal consistency of our approach on the high frame-rate Sintel dataset~\cite{Janai_CVPR_2017}.}\vspace{-0.3cm}
	\label{fig:sintel}
\end{figure}

\vspace{0.05in}
\noindent\textbf{Optical flow estimation.} To analyze how well our approach performs with different correspondence estimates, we consider three diverse state-of-the-art optical flow methods~\cite{Hui_CVPR_2018, Ilg_CVPR_2017, Sun_CVPR_2018}, each trained on FlyingChairs~\cite{Dosovitskiy_ICCV_2015}. As shown in Table~\ref{tbl:ablation} (fifth section), they all perform similarly well. Due to softmax splatting being fully differentiable, we are further able to fine-tune the optical flow estimation for the task of frame interpolation~\cite{Xue_IJCV_2019}. Specifically, we fine-tune PWC-Net and see additional improvements with this PWC-Net-ft that has been optimized for the task of frame interpolation. We thus use PWC-Net-ft for our approach.

\begin{figure*}\centering
	\setlength{\tabcolsep}{0.0cm}
	\renewcommand{\arraystretch}{1.2}
	\newcommand{\quantTit}[1]{\multicolumn{3}{c}{\scriptsize #1}}
	\newcommand{\quantSec}[1]{\scriptsize #1}
	\newcommand{\quantInd}[1]{\scriptsize #1}
	\newcommand{\quantVal}[1]{\scalebox{0.83}[1.0]{$ #1 $}}
	\newcommand{\quantBes}[1]{\scalebox{0.83}[1.0]{$\uline{ #1 }$}}
	\footnotesize
	\begin{tabularx}{\textwidth}{@{\hspace{0.1cm}} X P{1.9cm} P{1.08cm} @{\hspace{-0.31cm}} P{1.08cm} @{\hspace{-0.31cm}} P{1.08cm} P{1.08cm} @{\hspace{-0.31cm}} P{1.08cm} @{\hspace{-0.31cm}} P{1.08cm} P{1.08cm} @{\hspace{-0.31cm}} P{1.08cm} @{\hspace{-0.31cm}} P{1.08cm} P{1.08cm} @{\hspace{-0.31cm}} P{1.08cm} @{\hspace{-0.31cm}} P{1.08cm} P{1.08cm} @{\hspace{-0.31cm}} P{1.08cm} @{\hspace{-0.31cm}} P{1.08cm}}
		\toprule
			& & \quantTit{Vimeo-90k~\cite{Xue_IJCV_2019}} & \quantTit{Middlebury~\cite{Baker_IJCV_2011}} & \quantTit{UCF101 - DVF~\cite{Liu_ICCV_2017}} & \quantTit{Xiph - 2K} & \quantTit{Xiph - ``4K''}
		\\ \cmidrule(l{2pt}r{2pt}){3-5} \cmidrule(l{2pt}r{2pt}){6-8} \cmidrule(l{2pt}r{2pt}){9-11} \cmidrule(l{2pt}r{2pt}){12-14} \cmidrule(l{2pt}r{2pt}){15-17}
			& {\vspace{-0.37cm} \scriptsize training \linebreak dataset} & \quantSec{PSNR} \linebreak \quantInd{$\uparrow$} & \quantSec{SSIM} \linebreak \quantInd{$\uparrow$} & \quantSec{LPIPS} \linebreak \quantInd{$\downarrow$} & \quantSec{PSNR} \linebreak \quantInd{$\uparrow$} & \quantSec{SSIM} \linebreak \quantInd{$\uparrow$} & \quantSec{LPIPS} \linebreak \quantInd{$\downarrow$} & \quantSec{PSNR} \linebreak \quantInd{$\uparrow$} & \quantSec{SSIM} \linebreak \quantInd{$\uparrow$} & \quantSec{LPIPS} \linebreak \quantInd{$\downarrow$} & \quantSec{PSNR} \linebreak \quantInd{$\uparrow$} & \quantSec{SSIM} \linebreak \quantInd{$\uparrow$} & \quantSec{LPIPS} \linebreak \quantInd{$\downarrow$} & \quantSec{PSNR} \linebreak \quantInd{$\uparrow$} & \quantSec{SSIM} \linebreak \quantInd{$\uparrow$} & \quantSec{LPIPS} \linebreak \quantInd{$\downarrow$}
		\\ \midrule
SepConv - $\mathcal{L}_1$ & proprietary & \quantVal{33.80} & \quantVal{0.956} & \quantVal{0.027} & \quantVal{35.73} & \quantVal{0.959} & \quantVal{0.017} & \quantVal{34.79} & \quantVal{0.947} & \quantVal{0.029} & \quantVal{34.77} & \quantVal{0.929} & \quantVal{0.067} & \quantVal{32.06} & \quantVal{0.880} & \quantVal{0.169}
\\
SepConv - $\mathcal{L}_F$ & proprietary & \quantVal{33.45} & \quantVal{0.951} & \quantVal{0.019} & \quantVal{35.03} & \quantVal{0.954} & \quantVal{0.013} & \quantVal{34.69} & \quantVal{0.945} & \quantVal{0.024} & \quantVal{34.47} & \quantVal{0.921} & \quantVal{0.041} & \quantVal{31.68} & \quantVal{0.863} & \quantVal{0.097}
\\
ToFlow & Vimeo-90k & \quantVal{33.73} & \quantVal{0.952} & \quantVal{0.027} & \quantVal{35.29} & \quantVal{0.956} & \quantVal{0.024} & \quantVal{34.58} & \quantVal{0.947} & \quantVal{0.027} & \quantVal{33.93} & \quantVal{0.922} & \quantVal{0.061} & \quantVal{30.74} & \quantVal{0.856} & \quantVal{0.132}
\\
CyclicGen & UCF101 & \quantVal{32.10} & \quantVal{0.923} & \quantVal{0.058} & \quantVal{33.46} & \quantVal{0.931} & \quantVal{0.046} & \quantVal{35.11} & \quantVal{0.950} & \quantVal{0.030} & \quantVal{33.00} & \quantVal{0.901} & \quantVal{0.083} & \quantVal{30.26} & \quantVal{0.836} & \quantVal{0.142}
\\
CtxSyn - $\mathcal{L}_{\textit{Lap}}$ & proprietary & \quantVal{34.39} & \quantVal{0.961} & \quantVal{0.024} & \quantVal{36.93} & \quantVal{0.964} & \quantVal{0.016} & \quantVal{34.62} & \quantVal{0.949} & \quantVal{0.031} & \quantVal{35.71} & \quantVal{0.936} & \quantVal{0.073} & \quantVal{32.98} & \quantVal{0.890} & \quantVal{0.175}
\\
CtxSyn - $\mathcal{L}_F$ & proprietary & \quantVal{33.76} & \quantVal{0.955} & \quantVal{0.017} & \quantVal{35.95} & \quantVal{0.959} & \quantVal{0.013} & \quantVal{34.01} & \quantVal{0.941} & \quantVal{0.024} & \quantVal{35.16} & \quantVal{0.921} & \quantVal{0.035} & \quantVal{32.36} & \quantVal{0.857} & \quantVal{0.081}
\\
DAIN & Vimeo-90k & \quantVal{34.70} & \quantVal{0.964} & \quantVal{0.022} & \quantVal{36.70} & \quantVal{0.965} & \quantVal{0.017} & \quantVal{35.00} & \quantVal{0.950} & \quantVal{0.028} & \quantVal{35.95} & \quantVal{0.940} & \quantVal{0.084} & \quantVal{33.49} & \quantVal{0.895} & \quantVal{0.170}
\\
Ours - $\mathcal{L}_{\textit{Lap}}$ & Vimeo-90k & \quantBes{36.10} & \quantBes{0.970} & \quantVal{0.021} & \quantBes{38.42} & \quantBes{0.971} & \quantVal{0.016} & \quantBes{35.39} & \quantBes{0.952} & \quantVal{0.033} & \quantBes{36.62} & \quantBes{0.944} & \quantVal{0.107} & \quantBes{33.60} & \quantBes{0.901} & \quantVal{0.234}
\\
Ours - $\mathcal{L}_{\textit{F}}$ & Vimeo-90k & \quantVal{35.48} & \quantVal{0.964} & \quantBes{0.013} & \quantVal{37.55} & \quantVal{0.965} & \quantBes{0.008} & \quantVal{35.10} & \quantVal{0.948} & \quantBes{0.022} & \quantVal{35.74} & \quantVal{0.921} & \quantBes{0.029} & \quantVal{32.50} & \quantVal{0.856} & \quantBes{0.071}
		\\ \bottomrule
	\end{tabularx}\vspace{-0.1cm}
	\captionof{table}{Quantitative comparison of various state-of-the-art methods for video frame interpolation.}\vspace{-0.3cm}
	\label{tbl:comparison}
\end{figure*}

\vspace{0.05in}
\noindent\textbf{Perception-distortion tradeoff.} We train two versions of our model, one trained on color loss and one trained on perceptual loss, in order to account for the perception-distortion tradeoff~\cite{Blau_CVPR_2018}. As shown in Table~\ref{tbl:ablation} (sixth section), the model trained using color loss $\mathcal{L}_{\textit{Lap}}$ performs best in terms of PSNR and SSIM whereas the one trained using perceptual loss $\mathcal{L}_{\textit{F}}$ performs best in terms of LPIPS. We further note that the $\mathcal{L}_{\textit{F}}$-trained model better recovers fine details in challenging cases, making it preferable in practice.

\vspace{0.05in}
\noindent\textbf{Temporal consistency.} Since we use forward warping to compensate for motion, we can interpolate frames at an arbitrary temporal position despite only supervising our model at $t = 0.5$. To analyze the temporal consistency of this approach, we perform a benchmark on a high frame-rate version of the Sintel dataset~\cite{Janai_CVPR_2017}. Specifically, we interpolate frames $1$ through $31$ from frame $0$ and frame $32$ on all of its $13$ scenes. We include DAIN for reference since it is also able to interpolate frames at an arbitrary $t$. As shown in Figure~\ref{fig:sintel}, DAIN degrades around frame 8 and frame 24 whereas our approach via softmax splatting does not.

\subsection{Quantitative Evaluation}

We compare our approach to state-of-the-art frame interpolation methods on common datasets. Since these datasets are all low resolution, we also incorporate 4K video clips from Xiph which are commonly used to assess video compression. Specifically, we selected the eight 4K clips with the most amount of inter-frame motion and extracted the first $100$ frames from each clip. We then either resized the 4K frames to 2K or took a 2K center crop from them before interpolating the even frames from the odd ones. Since cropping preserves the inter-frame per-pixel motion, this ``4K'' approach allows us to approximate interpolating at 4K while actually interpolating at 2K instead. Directly processing 4K frames would have been unreasonable since DAIN, for example, already requires $16.7$ gigabytes of memory to process 2K frames. In comparison, our approach only requires $5.9$ gigabytes to process 2K frames which can be halved by using half-precision floating point operations.

As shown in Table~\ref{tbl:comparison}, our $\mathcal{L}_{\textit{Lap}}$-trained model outperforms all other methods in terms of PSNR and SSIM whereas our $\mathcal{L}_F$-trained model performs best in terms of LPIPS. Please note that on the Xiph dataset, all methods are subject to a significant degradation across all metrics when interpolating the ``4K'' frames instead of the ones that were resized to 2K. This shows that frame interpolation at high resolution remains a challenging problem. For completeness, we also show the per-clip metrics for the samples from Xiph in the supplementary material. We also submitted the results of our $\mathcal{L}_{\textit{Lap}}$-trained model to the Middlebury benchmark~\cite{Baker_IJCV_2011}. Our approach currently ranks first in this benchmark as shown in our supplementary material.

\subsection{Qualitative Evaluation} 

Since videos are at the heart of this work, we provide a qualitative comparison in the supplementary video. These support our quantitative evaluation and show difficult examples where our approach yields high-quality results whereas competing techniques are subject to artifacts.

\subsection{Discussion}

Our proposed softmax splatting enables us to extend and significantly improve the approach from Niklaus~\etal~\cite{Niklaus_CVPR_2018}. Specifically, softmax splatting enables end-to-end training which allows us to not only employ and optimize feature pyramids for image synthesis but also to fine-tune the optical flow estimator~\cite{Xue_IJCV_2019}. Our evaluation shows that these changes significantly improve the interpolation quality.

Another relevant approach is from Bao~\etal~\cite{Bao_CVPR_2019}. They forward-warp the optical flow and then backward warp the input images to the target location according to the warped optical flow. However, they use linear splatting and nearest neighbor interpolation. In comparison, our approach employs softmax splatting which is translational invariant and yields better results than linear splatting. Our approach is also conceptually simpler due to not warping the flow and not incorporating depth- or kernel-estimates. In spite of its simplicity, our approach compared favorably in the benchmark and, unlike DAIN, is temporally consistent.

The success of adversarial training as well as cycle consistency in image generation shows that more advanced supervision schemes can lead to improved synthesis results~\cite{Goodfellow_NIPS_2014, Liu_AAAI_2019, Reda_ICCV_2019, Zhu_ICCV_2017}. Such orthogonal developments could be used to further improve our approach in the future.

\section{Conclusion}
\label{sec:conclusion}

In this paper, we presented softmax splatting for differentiable forward warping and demonstrated its effectiveness on the application of frame interpolation. The key research question that softmax splatting addresses is how to handle cases where different source pixels forward-warp to the same target location in a differentiable way. Further, we show that feature pyramids can successfully be employed for high-quality image synthesis, which is an aspect of feature pyramids that has not been explored yet. Our proposed frame interpolation pipeline, which is enabled by softmax splatting and conceptually simple, compares favorably in benchmarks and achieves new state-of-the-art results.

\vspace{0.05in}
\noindent\textbf{Acknowledgments.} We are grateful for the feedback from Long Mai and Jon Barron, this paper would not exist without their support. All source image footage shown throughout this paper originates from the DAVIS challenge.

{\small
\bibliographystyle{ieee_fullname}
\bibliography{main}
}

\end{document}